\documentclass[letterpaper, 10 pt, conference]{ieeeconf}  

\IEEEoverridecommandlockouts                              

\usepackage{cite}
\usepackage{booktabs}
\usepackage{hyperref} 
\usepackage{multirow}
\usepackage{colortbl}
\usepackage{xcolor}
\usepackage{dblfloatfix}
\usepackage{xspace}
\usepackage{array}
\usepackage{amsmath} 
\usepackage{amssymb} 
\usepackage{bm}       
\usepackage{graphicx}
\usepackage{placeins} 
\usepackage{threeparttable}

\title{\LARGE \bf
Suite-IN: Aggregating Motion Features from Apple Suite for Robust Inertial Navigation
}

\author{Lan Sun$^{1*}$ 
\quad
Songpengcheng Xia $^{1*}$
\quad
Junyuan Deng$^{2}$
\quad
Jiarui Yang$^{1}$
\quad
Zengyuan Lai$^{1}$
\quad
Qi Wu$^{1}$
\quad
Ling Pei $^{1\dagger}$
\thanks{* indicates these authors contribute equally to this work.}
\thanks{$\dagger$ Indicates the corresponding authors}
\thanks{$^{1}$Lan Sun, Songpengcheng Xia, Jiarui Yang, Zengyuan Lai, Qi Wu, Ling Pei are with School of Electronic Information and Electrical Engineering, Shanghai Jiao Tong University, China. $^{2}$Junyuan Deng is with the Hong Kong University of Science and Technology. The work was supported by the National Natural Science Foundation of China under Grant 62273229.}
}

\begin{document}

\maketitle
\thispagestyle{empty}
\pagestyle{empty}

\begin{abstract}
With the rapid development of wearable technology, devices like smartphones, smartwatches, and headphones equipped with IMUs have become essential for applications such as pedestrian positioning. However, traditional pedestrian dead reckoning (PDR) methods struggle with diverse motion patterns, while recent data-driven approaches, though improving accuracy, often lack robustness due to reliance on a single device.
 In our work, we attempt to enhance the positioning performance using the low-cost commodity IMUs embedded in the wearable devices. We propose a multi-device deep learning framework named Suite-IN, aggregating motion data from Apple Suite for inertial navigation. Motion data captured by sensors on different body parts contains both local and global motion information, making it essential to reduce the negative effects of localized movements and extract global motion representations from multiple devices.
Our model innovatively introduces a contrastive learning module to disentangle motion-shared and motion-private latent representations, enhancing positioning accuracy. We validate our method on a self-collected dataset consisting of Apple Suite: iPhone, Apple Watch and Airpods, which supports a variety of movement patterns and flexible device configurations. Experimental results demonstrate that our approach outperforms state-of-the-art models while maintaining robustness across diverse sensor configurations.

\end{abstract}

\section{INTRODUCTION}
With the advancement of mobile computing and wearable technology, smart wearable devices like smartphones, smartwatches, and headphones equipped with inertial measurement units (IMUs) have become ubiquitous in daily life \cite{herath2020ronin, pan2024learning, jiang2024robust, chen2019motiontransformer}. IMUs (accelerometers, gyroscopes, and magnetometers) provide detailed motion data by being attached to various parts of the human body \cite{zhang2024dynamic, pei2021mars,9811933}. Research in areas like health monitoring, human-computer interaction, and augmented reality has widely leveraged these devices\cite{chen2019deep, mollyn2023imuposer, xia2024timestamp,10161112}.
\begin{figure}[t]
    \centering
    \includegraphics[width=\columnwidth]{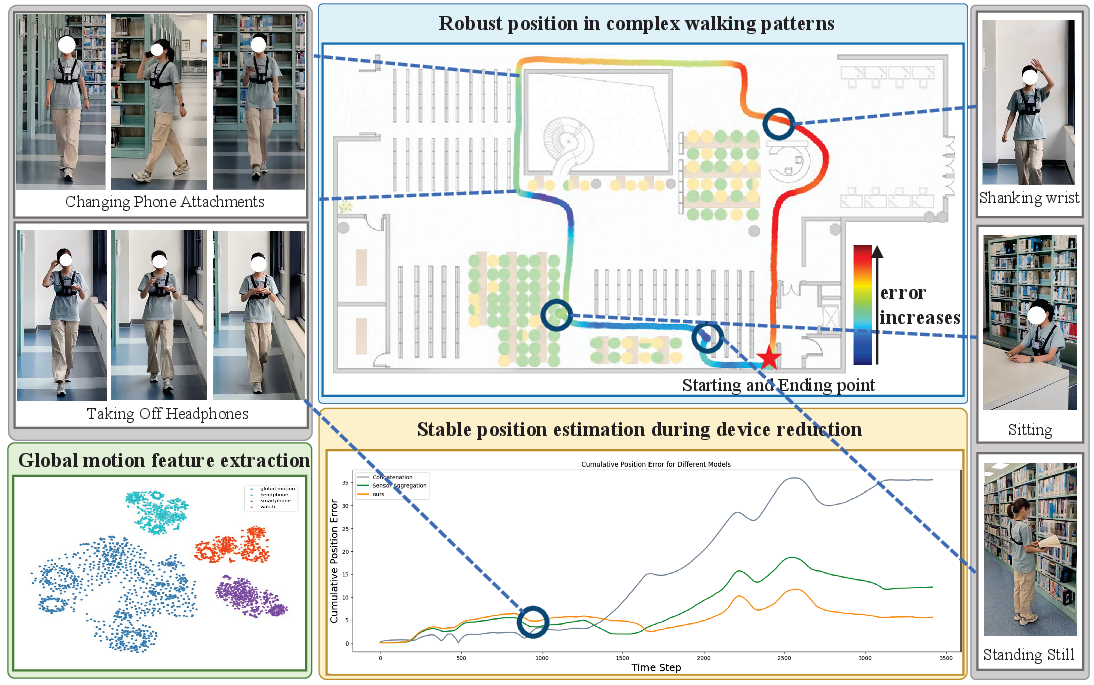} 
    \vspace{-5mm}
    \caption{Our innovative data-driven approach for robust pedestrian localization: achieving robust positioning under complex walking patterns and flexible equipment configurations by extracting global motion information.}
    \label{fig:1}
    \vspace{-5mm}
\end{figure}

Pedestrian dead reckoning (PDR) based on wearable devices, particularly smartphone, has gained attention for its ability to offer continuous and flexible positioning, especially in GNSS-denied environments \cite{herath2020ronin, pei2018optimal}. Early traditional PDR algorithms \cite{pei2018optimal} relied on multi-channel sensor signals like acceleration and angular velocity, coupled with motion constraints, to estimate movement direction and step count. However, these methods struggled to handle diverse motion patterns. Recent deep learning-based approaches \cite{herath2020ronin, chen2018ionet} have improved positioning accuracy by extracting high-level motion features like IONet \cite{chen2018ionet} and RONIN \cite{herath2020ronin}. Despite these advancements, relying on a single device limits robustness, making it challenging to accommodate the diverse motion patterns encountered in real-world environments.

As wearable devices have been more prevalent, it is a viable approach to leverage multiple wearable devices to form a sensor network that enhances positioning performance and address the challenges posed by diverse motion patterns. However, wearable devices are typically worn on different parts of the body, such as a watch on the wrist or headphones in the ears. The sensor data collected from different devices reflect the motion patterns of various body parts  \cite{pei2021mars}, which contain both global and local motion patterns, with local movements often introduce noise that may degrade positioning performance. The key challenge in multi-wearable fusion is effectively extracting global motion representations while minimizing the adverse impact of local noise.

\begin{table*}[t]
\centering
\caption{Comparison between Inertial Navigation Datasets.} 
\resizebox{\textwidth}{!}{  
\begin{tabular}{cccccc}  
\toprule
Dataset & Seqs & Sample Rate& Device & Device Flexibility & Walking Scenarios \\
\midrule
OxIOD   & 158 &100hz& iPhone & only change phone attachment   & small, medium \\
RoNIN   & 276 &200hz& Android phone & only change phone attachment   & medium \\
DeepIT  & /   &60hz & eSense, Android phone & flexible attachment but fixed device number            & small, medium, large \\
Suite-IN dataset   & 126 &25\textasciitilde100hz& iPhone, Watch, Airpods & both flexible attachment and device number  & small, medium, large \\
\bottomrule
\vspace{-2mm}
\end{tabular}
}
\begin{tablenotes}
\footnotesize
\item \textit{Small}: an area smaller than 30$\times$30 $m^2$, \textit{medium}: an area smaller than 50$\times$50 $m^2$, \textit{large}: an area larger than 100$\times$100 $m^2$.
\end{tablenotes}
\label{tab:1}
\vspace{-5mm}
\end{table*}
Our key insight is that wearable sensors on different body parts capture both global and local motion, but despite their placement or movement, they inherently share the same global motion information. To achieve accurate positioning, it is vital to separate global motion from localized movements to reduce the negative impact from local motion. We propose a multi-device deep learning framework that aggregates motion data from wearable sensors into a shared latent space. To further disentangle local movements, we use contrastive learning\cite{liu2024focal, cai2024orthogonality} to separate motion-shared and motion-private representations, enabling better aggregating global motion from sensors placed on different body parts.

\begin{figure}[t]
    \centering
    \includegraphics[width=0.5\textwidth]{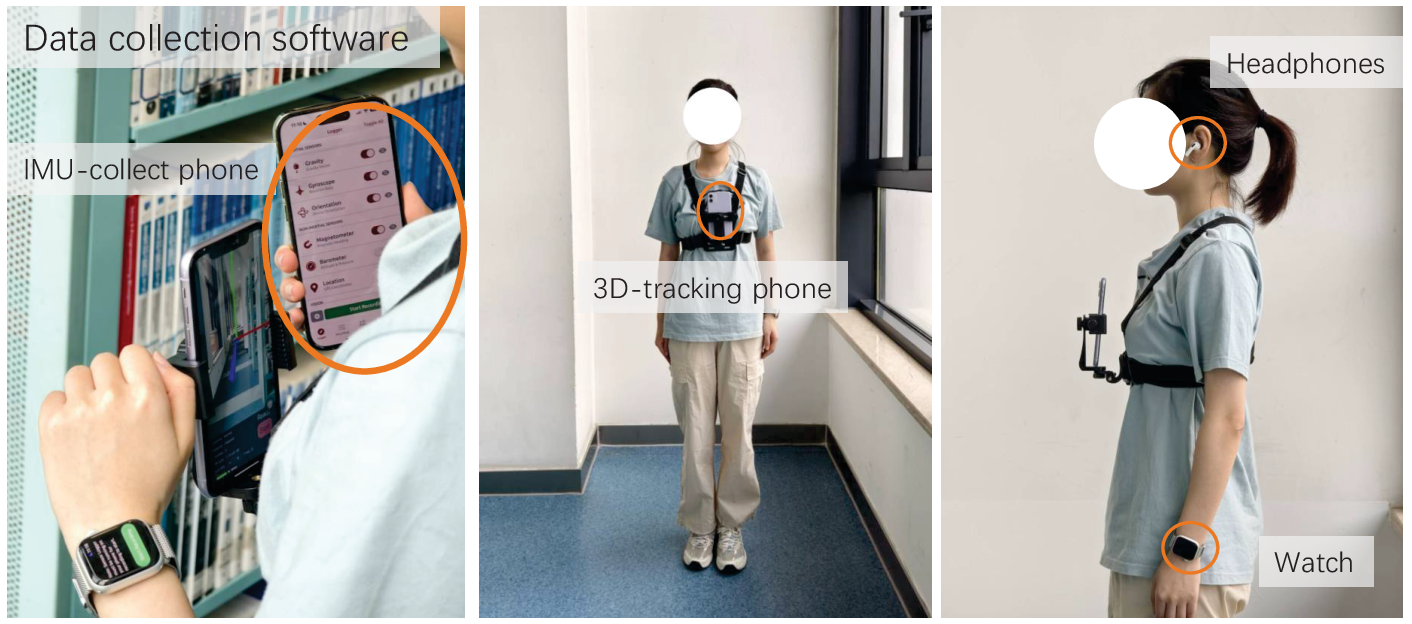} 
    \vspace{-5mm}
    \caption{Illustration of device wearing when we collect data}
    \label{fig:2}
    \vspace{-5mm}
\end{figure}

In addition, unlike previous work \cite{10447298, gong2021robust}, we use real-world wearable devices, Apple Suite (iPhone, Apple Watch and Airpods) for data collection. The consumer-grade IMUs in these devices tend to have higher noise levels compared to specialized IMU systems like Xsens \cite{schepers2018xsens} and Noitom \cite{noitom2019}. To ensure our approach is more applicable to real-life scenarios, we also allow for flexible smartphone holding manners and the configuration of the devices, as shown in Fig.\ref{fig:1}.
In summary, the key contributions of our paper are as follows:

\begin{itemize}

\item We introduce a novel deep learning framework that leverages multiple consumer-grade wearable devices, effectively fusing motion data from devices positioned across different body parts. This multi-device integration enables robust and accurate pedestrian positioning in diverse and dynamic environments.
\item To capture global motion while mitigating the effects of local body movements, we propose an innovative contrastive learning module. This module disentangles motion-shared and motion-aware latent representations, enhancing the model to isolate meaningful global motion features across varying body movements.
\item We first introduce a comprehensive pedestrian positioning dataset\footnote{https://github.com/LannnSun/Suite-IN-dataset} featuring data from smartphones, smartwatches, and headphones. It covers diverse motion patterns and flexible device configurations, offering a valuable resource for multi-device positioning research. Experiments show that our method outperforms state-of-the-art approaches, providing robust and accurate pedestrian localization in real-world scenarios.
\end{itemize}



\section{Related Work}
In this section, we review several related works on these two topics: data-driven pedestrian localization and contrastive learning for wearable sensors.

\subsection{Data-driven Pedestrian Localization}

Data-driven smartphone inertial odometry has gained significant attention recently. IONet \cite{chen2018ionet} uses a long short-term memory (LSTM) network to estimate pedestrian velocity and heading changes from smartphone data. RIDI \cite{yan2018ridi} classifies phone attachment using a support vector machine, followed by velocity regression for each attachment. RoNIN \cite{herath2020ronin} employs LSTM, ResNet, and TCN models to predict pedestrian velocity. To handle domain differences in inertial data, MotionTransformer \cite{chen2019motiontransformer} leverages GAN and domain adaptation for improved navigation. DeepIT \cite{gong2021robust} fuses IMU data from smartphones and earbuds using a reliability network for inertial tracking. Song et al. \cite{10447298} introduce Restrained-Weighted-Fusion to improve the accuracy and robustness of multi-node fusion positioning.

\subsection{Contrastive Learning for Wearable sensors}
Contrastive learning \cite{deldari2022cocoa,liu2024focal,fortes2022learning,xia2024timestamp} has been extensively studied and applied in wearable-sensor-based human-centric-tasks,such as human activity classification(HAR). 
The main idea of contrastive learning is to learn fair representation in a discriminative manner by using InfoNCE loss or its variants\cite{deldari2022cocoa}.
COCOA\cite{deldari2022cocoa} employs contrastive learning to learn quality representations from multisensor data by computing the cross-correlation between different data modalities. 
Considering that the modality-specific representations also play an important role in the downstream task, Liu et al.\cite{liu2024focal} use an orthogonality restriction and simultaneously leverage the modality-shared and modality-specific representations based on contrastive loss.

\begin{figure*}[t]
    \centering
    \includegraphics[width=\textwidth]{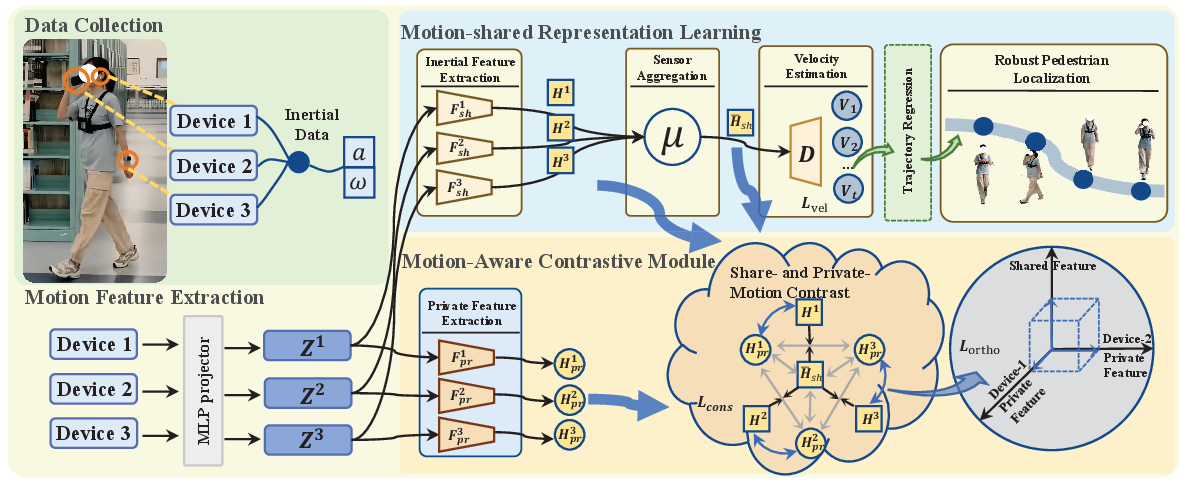} 
    \caption{Overview of our multi-device fusion positioning framework. Our proposed model comprises two technical modules: 1) motion-shared representation learning and 2) motion-aware contrastive learning module. } 
    \label{3}
    \vspace{-5mm}
\end{figure*}

\section{Dataset Description}

\subsection{Dataset overview}
Our dataset comprises 126 sequences, totaling 6.5 hours of data and covering 18 km of walking, collected from 3 subjects in 3 different scenes. It includes various device attachments (handheld, pocket, bag) and daily walking patterns (e.g., removing a device while walking, sitting, or standing still) to simulate real-world conditions. Compared to mainstream datasets like OxIOD \cite{chen2020deep} and RoNIN \cite{herath2020ronin}, our dataset spans a wider range of walking scenarios and complex patterns. Detailed information is shown in Table \ref{tab:1}. The dataset's diversity supports not only positioning tasks but also applications like Human Action Recognition (HAR)\cite{9535462,10001226} and Human Pose Estimation (HPE)\cite{huang2018deep,dai2023sloper4d}.
\subsection{Data Collection Apparatus and Process}
Our data collection setup included an Apple iPhone 14 Pro, an Apple Watch Series 8 worn on the left wrist, and a pair of Apple AirPods 3. The iPhone and Watch both sampled at 100 Hz, while the AirPods sampled at 25 Hz. To ensure uniformity, the system’s overall sampling rate is set to 25 Hz, the maximum rate of the AirPods. The Apple Watch and AirPods communicated with the iPhone via Bluetooth, and the iPhone transmits all IMU data to a laptop for processing.
We used Apple ARKit\footnote{https://developer.apple.com/documentation/arkit/}, Apple’s augmented reality (AR) framework, which includes a tightly coupled VIO (visual-inertial odometry) algorithm, allowing iOS devices to track movement in 3D space. The relative pose error from Apple ARKit is about 0.02 m of drift error per second\cite{kim2022benchmark}. To obtain ground truth positions, we attach a 3D tracking phone (iPhone 11) to the body using a harness, while subjects handled the other phone freely for IMU data collection. The 3D tracking phone saved pose estimates as a translation vector and unit quaternion at 30 Hz. An illustration of the device setup is shown in Fig.\ref{fig:2}. Before data collection, we calibrate the IMU sensor biases and aligned the VIO and IMU systems temporally and spatially.

\section{Method}

\subsection{Problem Statement}
\label{PS}
In this article, we define the $D$-dimensional wearable sensory sequence of length $T$ as $\mathbf{X_{1:T}} = [\mathbf{x}_1,...,\mathbf{x}_T ], \mathbf{x}_t \in \mathbb{R}^{D}$. With $J$ wearable devices in our dataset, $\mathbf{x}_t$ could be denoted as $\mathbf{x}_t = [\mathbf{x}^1_t,...,\mathbf{x}^J_t], \mathbf{x}^j_t = [\mathbf{a}\ \boldsymbol{\omega}] \in \mathbb{R}^{6}$, which contains the acceleration and angular velocity for $j$-th wearable device at the $t$-th time step. For the pedestrian localization, the sequence is divided into windows of length $L$ using a sliding window approach. Each window $n \in \{1,...,n,...,N\}$ is denoted as $\mathbf{X}_n = [\mathbf{x}_t,...,\mathbf{x}_{t+L-1}]$.

Our goal is to predict the mean velocity $\mathbf{v}_n=[\mathbf{v}_x\ \mathbf{v}_y] \in \mathbb{R}^{2}$ within each window. By integrating the estimated velocities $\mathbf{v}_n, n \in \{1,...,N\}$, we can ultimately accumulate and obtain the user's trajectory. Additionally, each sensor modality generates intermediate features—motion-shared features $\mathbf{H}^{j}$ and private features $\mathbf{H}^{j}_{pr}$, which are used in contrastive learning to improve localization accuracy. As shown in Fig.\ref{3}, our proposed model comprises two technical modules: 1) motion-shared representation learning for pedestrian localization estimation and 2) motion-aware contrastive learning module for shared and private space.

\subsection{Motion-shared Representation Learning for pedestrian localization estimation}
Wearable devices are often placed on different parts of the body, leading to complex and irregular motion data due to limb movements. However, the motion information related to the user's trajectory is embedded in the latent common representations across various devices. To address the heterogeneity in motion perception caused by the flexible device placement, we designed a motion-shared representation learning module that fuses heterogeneous sensor data into a shared low-dimensional space, obtaining robust motion-shared representations for localization. Therefore, this module comprises three sub-modules: 1) Independent wearable sensor feature extraction, 2) Sensor feature aggregation, 3) Velocity and trajectory regression.

\textbf{Independent wearable sensor feature extraction:}
As described in Fig.\ref{3}, the sensor data $\mathbf{X} = [\mathbf{X}^1,...,\mathbf{X}^J]$ from $J$ wearable devices is processed with a multilayer perceptron (MLP) network $F_{MLP}(\cdot)$ to extract shallow features $\mathbf{Z}$. These features are then divided by device, resulting in independent shallow features $\mathbf{Z}=[\mathbf{Z}^1,...,\mathbf{Z}^J]$. Independent feature extractors $F^j_{sh}(\cdot)$ are applied to each shallow feature $\mathbf{Z}^J$, extracting the their own motion features $\mathbf{H}^j$ as follows: 
\begin{equation}
\mathbf{H}^j = F^j_{sh}(\mathbf{Z}^j).
\end{equation}

Each independent feature extractors $F^j_{sh}(\cdot)$ is comprised of two components. First, a two-layer 2D convolutional network with max pooling extracts motion features across both temporal and sensor channel dimensions. The kernel size is 1 along the sensor channel and 3 along the temporal dimension, allowing the convolution to capture temporal patterns without affecting sensor dimensionality \cite{pei2021mars}. The sensor and convolutional channels are then fused into higher-dimensional features. A time-distributed dense layer \cite{jeyakumar2019sensehar} is applied at each time step, maintaining the sensor channel dimensionality for efficient feature aggregation in later stages.


\textbf{Sensor feature aggregation:}
After extracting motion features $\mathbf{H}^j$ from each wearable device, we aggregate them to capture the global motion information. Although different devices are worn on various body parts to collect motion data, they all follow a common principle: each wearable device contains global motion features, regardless of its placement or the different motion patterns. We consider this to be the shared component of each wearable device's motion representation. Inspired by \cite{jeyakumar2019sensehar}, we adopt a simple yet effective approach, taking the arithmetic mean ($\mu$) of the motion features from all devices. The aggregated shared feature, denoted as $\mathbf{\bar{H}}_{sh}$, fuses the heterogeneous sensor data into a shared low-dimensional latent space that better represents the global motion features.

\textbf{Velocity and trajectory regression:}
Following previous works\cite{10447298, herath2020ronin}, we take the average velocity of the window as the network's output with a velocity regression network $D(\cdot)$, denoted as $\hat{\mathbf{v}} = D(\mathbf{\bar{H}}_{sh})$
To ensure comprehensive feature extraction, velocity is also regressed for each device’s features, yielding $\hat{\mathbf{v}}^j = D( \mathbf{H}^j)$. The velocity loss is calculated using Mean-Squared-Error (MSE) loss: 
\begin{equation}
   \mathcal{L}_{vel}=\frac{1}{J+1} \sum_{j=0}^{P}MSE(\hat{\mathbf{v}}^j,\mathbf{v}).
   \label{lvel}
\end{equation}
For convenience, we denote the aggregated features $\mathbf{\bar{H}}_{sh}$ as $\mathbf{{H}}^0$, regarding it as the $0$-th sensor. The velocity regressed from the aggregated features is denoted as $\hat{\mathbf{v}}^0$. 
However, our ultimate goal is to obtain the pedestrian's motion trajectory $\hat{\mathbf{y}}_t$. We update the velocity $\hat{\mathbf{v}}_t$ at each sampling moment using the predicted average velocity regressed from the aggregated features, and then integrate over time to obtain the pedestrian's trajectory,
 \begin{equation}
\hat{\mathbf{y}}_t =\mathbf{y}_{t_0} +\int_{t_0}^{t} \hat{\mathbf{v}}_t dt.
\end{equation}

\subsection{Motion-aware Contrastive Module for Share- and Private-motion Learning}
Based on our insights, human motion contains both global and local motion information. 
Wearable devices, placed on different body parts, capture varyious local motion data, which can negatively impact positioning tasks. While our method attempts to extract global motion features through feature aggregation, it doesn’t fully separate local from global motion information. Therefore, we designed a motion-aware contrastive learning module to disentangle local features from shared global features. Shared features capture global motion, while private features represent local motion from different sensors.
To achieve this separation, the model additionally estimates private features alongside the shared ones. We use a private feature extractor $F^j_{pr}(\cdot)$ for each wearable device, extracting private features $\mathbf{H}^j_{pr}$ from the shallow feature $\mathbf{Z}^j$. The previously extracted motion representation $\mathbf{H}^j$ serves as the shared feature. Together with the private features, we apply contrastive learning loss and orthogonality constraints to structure the feature space effectively \cite{tian2020contrastive, liu2024focal}.

\textbf{Contrastive learning for share-motion representation:}
Our model iterates over all extracted features, treating the aggregated shared feature and each modality's shared feature $(\mathbf{\bar{H}}_{sh},\mathbf{H}^j)$ as positive pairs. In contrast, the aggregated shared feature and each modality's private feature $(\mathbf{\bar{H}}_{sh},\mathbf{H}^j_{pr})$, as well as the private features between different modalities $(\mathbf{H}^i_{pr},\mathbf{H}^j_{pr})$, are treated as negative pairs. With these positive and negative pairs, we compute the contrastive loss using InfoNCE \cite{tian2020contrastive, liu2024focal} as follows:
\begin{equation}
\left\{  
             \begin{array}{lr}  
             \small
 \mathcal{L}_{con}=-\sum_{j=1}^{J}log
\frac{s\left ( \mathbf{\bar{H}}_{sh},\mathbf{H}^j \right ) }{s\left ( \mathbf{\bar{H}}_{sh},\mathbf{H}^j \right )+ S\left ( \mathbf{\bar{H}}_{sh},\mathbf{H}^j_{pr} \right ) +S\left ( \mathbf{H}^i_{pr},\mathbf{H}^j_{pr} \right )} &  \\  
             \left ( \mathbf{\bar{H}}_{sh},\mathbf{H}^j \right )  =exp\left (   \left \langle  \mathbf{\bar{H}}_{sh},\mathbf{H}^j \right \rangle /\tau\right ) & \\  
             \left (\mathbf{\bar{H}}_{sh},\mathbf{H}^j_{pr}  \right )= \sum_{j=1}^{P} exp\left ( \left \langle \mathbf{\bar{H}}_{sh},\mathbf{H}^j_{pr}  \right \rangle /\tau \right ) & \\
 \left ( \mathbf{H}^i_{pr},\mathbf{H}^j_{pr} \right )= \sum_{i,j\in [0,...,J],i \ne j}exp\left ( \left \langle  \mathbf{H}^i_{pr},\mathbf{H}^j_{pr}\right \rangle /\tau  \right ),& \\
             \end{array}  
\right. 
\label{lcon}
\end{equation}
where $\left \langle \cdot  \right \rangle $ means calculating cosine similarity.

\textbf{Orthogonality Constraint for Share- and Private-motion Representation:}
To prevent shared global motion information from being reused in the private space and to capture unique local motion data from different wearable devices, we apply orthogonality constraints inspired by \cite{liu2024focal, cai2024orthogonality}. We apply the cosine embedding loss\cite{techapanurak2020hyperparameter} between shared and private features, as well as between the private features themselves, which can be expressed as:
\begin{equation}
\mathcal{L}_{orth}=\sum_{i,j\in [0,...,J],i \ne j} \left \langle \mathbf{H}^i_{pr},\mathbf{H}^j_{pr} \right \rangle +\sum_{j=1}^{J}\left \langle \mathbf{H}^j,\mathbf{H}^j_{pr} \right \rangle  .
\label{lorth}
\end{equation}

\begin{table*}[t]
\centering
\caption{Positioning evaluation of various methods under different walking modes}
\scalebox{1.2}{
\begin{tabular}{|c|c|c|c|c|c|c|c|c|}
\hline
 & \textbf{Metric} &\textbf{Overall} & \textbf{SA} & \textbf{SS} &\textbf{TOH} &\textbf{TOW} & \textbf{ATOW} &\textbf{ATOH} \\ \hline

\multirow{2}{*}{\textbf{IONet*\cite{chen2018ionet}}}  & \textit{ATE}                    & 6.080   & 9.268  & \underline{5.704}  & 5.382  & 5.482  & 12.746 & 9.929  \\ \cline{2-9} 
                        & \textit{RTE}                    & 7.264   & 11.311 & \underline{5.713}  & 7.959  & 7.894  & 15.616 & 10.751 \\ \hline
\multirow{2}{*}{\textbf{RoNIN*\cite{herath2020ronin}}}  & \textit{ATE}                    & \underline{5.110}   & 10.225 & 5.854  & 6.373  & 6.337  & \underline{8.037}  & \underline{5.495}  \\ \cline{2-9} 
                        & \textit{RTE}                    & \underline{6.793}   & 13.377 & 6.799  & 9.009  & 8.617  & \underline{10.769} & \underline{6.314}  \\ \hline
\multirow{2}{*}{\textbf{DeepIT\cite{gong2021robust}}} & \textit{ATE}                    & 11.760  & 14.478 & 12.419 & 10.897 & 10.994 & 12.957 & 17.835 \\ \cline{2-9} 
                        & \textit{RTE}                    & 14.272  & 17.371 & 13.407 & 15.799 & 15.693 & 22.203 & 17.502 \\ \hline
\multirow{2}{*}{\textbf{ReWF1\cite{10447298}}}  & \textit{ATE}                    & 5.918   & 12.247 & 4.715  & \underline{4.560}  & \underline{4.840}  & 8.663  & 11.753 \\ \cline{2-9} 
                        & \textit{RTE}                    & 7.656   & 10.853 & 7.548  & \underline{6.860}  & \underline{7.857}  & 11.221 & 13.057 \\ \hline
\multirow{2}{*}{\textbf{Suite-IN}}   &\textit{ATE}                     & \textbf{3.402}   & \textbf{4.259}  & \textbf{2.876}  & \textbf{3.785}  & \textbf{2.465}  & \textbf{3.192}  & \textbf{4.248}  \\ \cline{2-9} 
                        & \textit{RTE}                    & \textbf{4.660}   & \textbf{3.496}  & \textbf{3.450}  & \textbf{6.126}  & \textbf{4.086}  & \textbf{3.890}  & \textbf{3.817}  \\ \hline
\end{tabular}}
\begin{tablenotes}
\footnotesize
\item SA: Shake All devices randomly during walking. SS: Sit down and Stand still randomly. TOH: simulated Take Off Headphones. TOW: simulated Take Off Watch. ATOW: Actual Take Off Watch. ATOH: Actual Take Off Headphones. 
\end{tablenotes}
\vspace{-3mm}
\label{tab:2}

\end{table*}
\begin{figure*}[ht]
    \centering
    \includegraphics[width=0.8\textwidth]{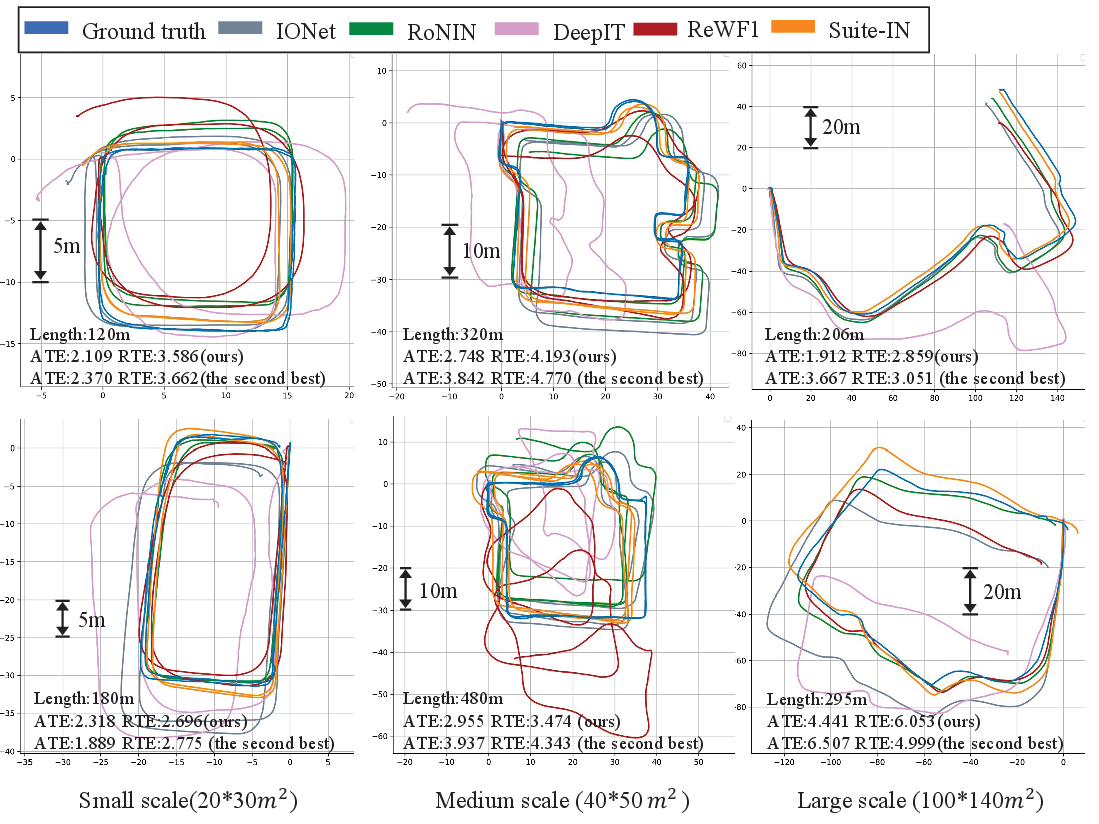} 
    \vspace{-4mm}
    \caption{Selected visualizations. We select 2 examples from each walking range and visualize the reconstructed trajectories of competing methods. For each sequence, we label the trajectory length, ATE and RTE of our method and the second-best method. }
    \label{fig:4}
    \vspace{-5mm}
\end{figure*}
\subsection{Overall Network Parameters Optimization}
With above loss functions introduced in Eq.\ref{lvel}, Eq.\ref{lcon} and Eq.\ref{lorth}, we set the $\lambda_v$, $\lambda_c$ and $\lambda_o$ as hyper-parameters that determine different loss’s contribution and obtain the final loss function,
\begin{equation}
\min \mathcal{L}=\lambda_v\cdot\mathcal{L}_{vel}+\lambda_c\cdot \mathcal{L}_{con}+\lambda_o\cdot \mathcal{L}_{orth}.
\end{equation}
We set the hyper-parameters$\lambda_v=1$, $\lambda_c=0.2$, $\lambda_o=0.05$.

\section{Experiment}

\subsection{Experimental Setup}

\textbf{(1)Evaluation Metrics}

 We employ two standard metrics from \cite{zhang2018tutorial}, to evaluate our results:\textbf{1)Absolute Trajectory Error (ATE)}, which measures the cumulative error between the predicted and reference trajectories using Root Mean Squared Error (RMSE), and \textbf{2)Relative Trajectory Error (RTE)} defined as the average RMSE between the predicted and reference trajectories over fixed time intervals.

 \textbf{(2)Competing Methods}
\label{CM}

\textbf{IONet*\cite{chen2018ionet}:}A deep-learning-based inertial navigation method using a LSTM network model. We realize IONet's three-node positioning by concatenating the data of three devices, denoted as IONet*.

\textbf{RoNIN*\cite{herath2020ronin}:} A deep-learning-based method for inertial tracking, which uses three different backbones (LSTM, Resnet, TCN). Resnet has the highest accuracy, and we combine the official implementation of RoNIN (Resnet) with concatenation approach, denoted as RoNIN*.

\textbf{DeepIT\cite{gong2021robust}:}An inertial navigation method integrating smartphone and headphones based on LSTM. We extend it to three-sensor fusion with primal weighting.

\textbf{ReWF1\cite{10447298}:}Three-node inertial positioning method consisted of Resnet-based inertial encoder and LSTM-based sensor weight extractor. We locally implement ReWF1 algorithm as the code is not public.

 \begin{figure*}[t]
    \centering
    \includegraphics[width=\textwidth]{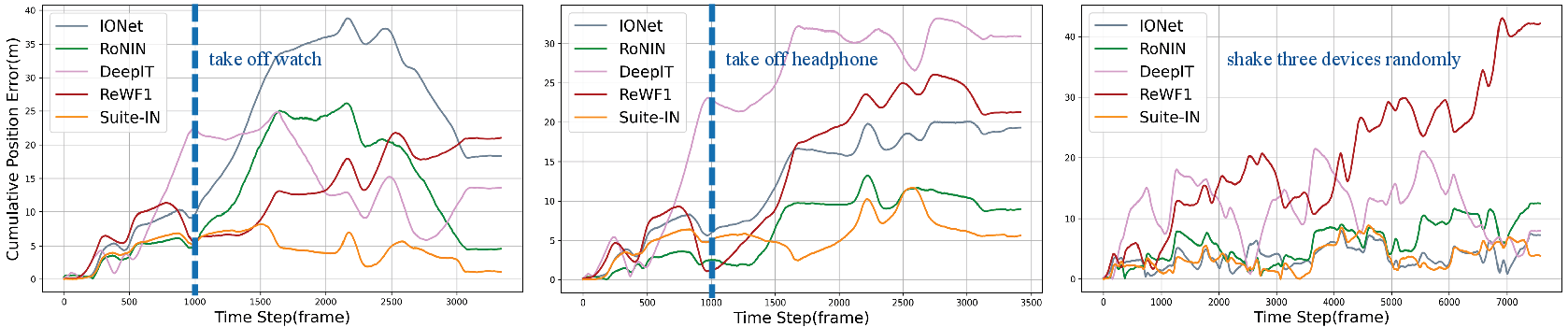} 
    \vspace{-7mm}
    \caption{Position Estimation Error. Left: the position estimation error of the sequence where the watch is taken off midway.  Middle: the sequence where the headphones are taken off midway. Right: the sequence where three devices are randomly shaken.}
    \label{fig:5}
    \vspace{-5mm}
\end{figure*}

\subsection{ Comparison with the State-of-the-Art}
For fair and meaningful evaluation, we trained all competing models on the same setting, and compared their performance to our model. Tab.\ref{tab:2} summarizes the localization performance of various competing algorithms.
\begin{table}[t]
\centering
\caption{Multi-sensor fusion effectiveness verification. }
\vspace{-2mm}
\resizebox{0.49\textwidth}{!}{
\begin{tabular}{|c|c|c|c|c|c|c|}
\hline

                  & \textbf{Metric} &\textbf{Overall}& \textbf{MP} & \textbf{HD} & \textbf{PK} & \textbf{BG} \\ \hline
\multirow{2}{*}{\textbf{IONet\cite{chen2018ionet}}} & \textit{ATE} & 13.542 &3.559 & 4.358 & 6.438 & 37.815 \\ \cline{2-7} 
                  & \textit{RTE} & 11.304 & 5.240 & 4.486 & 6.570 & 27.706 \\ \hline
\multirow{2}{*}{\textbf{RoNIN\cite{herath2020ronin}}} & \textit{ATE} & 8.109 & 5.152 & \textbf{3.736} & 11.435 & 11.522 \\ \cline{2-7} 
                  & \textit{RTE} & 7.927 & 4.886 & \textbf{4.085} & 10.814 & 11.317 \\ \hline
\multirow{2}{*}{\textbf{Suite-IN}} & \textit{ATE} & \textbf{4.414} & \textbf{3.457} & 4.035 & \textbf{3.496} & \textbf{6.478} \\ \cline{2-7} 
                  & \textit{RTE} & \textbf{4.553} & \textbf{3.947} & 5.300 & \textbf{3.166} & \textbf{5.678} \\ \hline
\end{tabular}}
\begin{tablenotes}
\footnotesize
\item IONet and RoNIN are implemented based on data from smartphone.
\end{tablenotes}
\label{tab:3}
\vspace{-6mm}
\end{table}
\textbf{Overall Performance Comparison: }In Tab.\ref{tab:2}, we use abbreviations such as \textbf{SA} and \textbf{SS} to represent different walking patterns and device configurations in our multi-device inertial dataset. The evaluation results clearly show the superior performance of our method compared to competing approaches across various walking patterns.
For particularly challenging scenarios such as \textbf{SS}, \textbf{ATOW}, and \textbf{ATOH}, our algorithm significantly reduced ATE and RTE. Specifically, ATE and RTE decreased by 2.83m (49.58\%) and 2.26m (39.61\%) for SS, 4.85m (60.28\%) and 6.88m (63.99\%) for ATOW, and 1.25m (22.69\%) and 2.50m (39.55\%) for ATOH, when compared to the second-best algorithm. Fig.\ref{fig:4} provides visual comparisons of the trajectories reconstructed by different methods alongside the ground truth. Our method outperforms competing approaches on various walking range.  

\textbf{Performance visualization for flexible device configuration:} Fig.\ref{fig:5} shows the positioning performance of our method compared to others in scenarios that support flexible device configuration, which is common in real-world applications. Our algorithm supports flexible device removal, with no abnormal fluctuations in position estimation error when the watch or headphones are removed during walking, as shown in Fig.\ref{fig:5}. In contrast, these competing methods show significant increases in estimation error under the same conditions. When the three devices are randomly shaken during walking, our algorithm maintains a consistently lower error than the comparison methods.


\textbf{Comparison of single device SOTA methods}
Tab.\ref{tab:3} compares the performance of two SOTA smartphone-based position methods, IONet and RoNIN, which rely solely on smartphone data. Along with overall positioning performance, we report results for different phone holding variations: \textbf{MP} (Multi-changed Phone hold manner), \textbf{HD} (phone HandhelD), \textbf{PK} (phone in PocKet), and \textbf{BG} (phone in BaG). Notably, \textbf{PK} and \textbf{BG} modes were excluded from the training set to assess generalization ability. While RoNIN and IONet perform well in \textbf{MP} and \textbf{HD} modes, their accuracy declines significantly in unseen \textbf{PK} and \textbf{BG} modes, as shown in Tab.\ref{tab:3}.
However, our method, leveraging multiple wearable devices, delivers more robust and stable positioning. Even in unseen \textbf{PK} and \textbf{BG} modes, our model benefits from the additional wearable sensor data, resulting in superior generalization and consistently accurate positioning.

In summary, our approach improves robustness over single-device methods by leveraging multi-device motion data and adapts to various sensor configurations for versatile real-world applications.


\begin{table}[t]
\centering
\caption{The ablation study on feature aggregation and contrastive learning based motion learning.}
\vspace{-2mm}
\resizebox{0.49\textwidth}{!}{
\begin{tabular}{|c|c|c c|c c|}
\hline
\textbf{Experiment} &\textbf{Method} & \textbf{MS.Module} & \textbf{MA.Module} & \textbf{ATE} & \textbf{RTE} \\
\hline

\multirow{3}{*}{1} &\multirow{3}{*}{Suite-IN}  &  &  & 5.941 & 7.769 \\

        &    & \checkmark &  & 4.489 & 5.822 \\

& & \checkmark & \checkmark & \textbf{3.402} & \textbf{4.660} \\

\hline
\multirow{4}{*}{2} &\multicolumn{3}{c|}{IONet*} & 6.080 & 7.264 \\

& \multicolumn{3}{c|}{IONet* with MS.Module} & 5.789 & 6.792 \\
\cline{2-4}
&\multicolumn{3}{c|}{RoNIN*} & 5.110 & 6.723 \\

&\multicolumn{3}{c|}{RoNIN* with MS.Module}  & 4.489 & 5.822 \\
\hline
\end{tabular}}
\label{tab:4}
\vspace{-5mm}
\end{table}

\subsection{Ablation study}

We further demonstrate the effectiveness of Motion-Shared representation learning (MS.Module) and Motion-Aware contrastive learning Module (MA.Module) in our method. The results in Table \ref{tab:4} shows a clear increase in ATE and RTE as each component is progressively removed, highlighting the importance of both modules for robust and accurate positioning. Additionally, we integrated MS.Module into the multi-sensor versions of IONet and RoNIN. As shown in Table \ref{tab:4}, our module consistently improves performance across different methods.

\section{CONCLUSIONS}
This study introduces a multi-device inertial dataset with flexible device configurations and presents a novel inertial positioning method based on wearable devices placed on different parts of the body. By integrating sensor data through motion-shared representation learning and motion-aware contrastive learning modules, our approach achieves robust and accurate positioning results. Importantly, our method supports the removal or reconfiguration of devices without significant performance degradation, making it highly practical for real-world applications. 
For localization tasks, the ability to distinguish between shared and private motion representations reduces the negative impact of local movements, thereby improving overall positioning accuracy. Additionally, the separated private features, which capture localized body motion information, hold potential for supporting other motion analysis tasks.



\bibliographystyle{ieeetr}
\bibliography{reference}
\addtolength{\textheight}{-12cm}   




\end{document}